\documentclass{article}
\usepackage[T1]{fontenc}
\usepackage{spconf,amsmath,amssymb,booktabs,graphicx}
\usepackage[hidelinks]{hyperref}
\usepackage{balance}
\usepackage{flafter}
\usepackage{orcidlink}

\graphicspath{{figures/}}

\renewcommand{\arraystretch}{0.95}
\newcommand{\sg}{\operatorname{sg}}
\newcommand{\Lcross}{\mathcal{L}_{\mathrm{cross}}}
\newcommand{\Lfut}{\mathcal{L}_{\mathrm{fut}}}
\newcommand{\Lspec}{\mathcal{L}_{\mathrm{spec}}}
\newcommand{\Lvar}{\mathcal{L}_{\mathrm{var}}}
\newcommand{\Lcov}{\mathcal{L}_{\mathrm{cov}}}

\title{Phenomenon-Graph JEPA: Label-Efficient Representation Learning\\
for Contactless Cardiorespiratory Sensing}

\name{\begin{tabular}{c}
Constantino Álvarez Casado$^{\star\dagger}$$^{\orcidlink{0000-0002-3052-4759}}$, Nhi Nguyen$^{\star}$$^{\orcidlink{0009-0002-2090-0746}}$, Mohammad Rakibur Rahman$^{\star}$$^{\orcidlink{0009-0003-8994-1033}}$, Le Nguyen$^{\star}$$^{\orcidlink{0000-0001-7765-1483}}$,\\
Manuel Lage Cañellas$^{\star\dagger}$$^{\orcidlink{0000-0002-4917-340X}}$, Sasan Sharifipour$^{\star}$$^{\orcidlink{0000-0001-6670-4252}}$, Miguel Bordallo López$^{\star}$$^{\orcidlink{0000-0002-5707-9085}}$
\end{tabular}}

\address{$^{\star}$Center for Machine Vision and Signal Analysis (CMVS), University of Oulu, Finland\\
$^{\dagger}$Candour Ltd, Oulu, Finland}

\begin{document}
\ninept
\maketitle

{\let\thefootnote\relax\footnotetext{This work has been submitted to the IEEE for possible publication. Copyright may be transferred without notice, after which this version may no longer be accessible. This extended version adds methodological details, a protocol schematic, implementation settings and further analyses to the conference submission.}}

\begin{abstract}
Millimeter-wave (mmWave) radar and RGB-D cameras can record cardiac and respiratory waveforms continuously and without contact, but labeled recordings remain scarce because every label requires a supervised acquisition session. Self-supervised pretraining can exploit the unlabeled signals, yet contrastive methods depend on signal transformations and negative pairs whose validity is uncertain for cardiorespiratory data, where time warping changes breathing rate and distant windows can share the same physiological state. We present Phenomenon-Graph JEPA, a joint-embedding predictive architecture that learns from four processed one-dimensional streams without negative pairs or synthetic augmentation in its base configuration. Each stream is encoded by a temporal convolutional branch and a band-limited spectral branch. During pretraining, the model predicts stopped target embeddings along typed edges, which connect streams assigned to the same physiological phenomenon, and forward in time within a state episode. We treat this physiological typing as a testable hypothesis and compare it with wrong-edge and all-pairs prediction graphs. In the OMuSense-23 dataset, pretraining improves label-efficiency area over matched supervised training by 3.91 percentage points (95\% interval 2.08 to 5.80, Holm-adjusted $p=0.006$), and by 3.74 points under a second configuration evaluated on the same test participants. However, the wrong-edge and all-pairs controls do not establish a benefit from physiological typing. Optional Takens-inspired delay coordinates improve a validation comparison with learned history, whereas two wrist-only WESAD protocols do not establish a pretraining advantage. The study therefore separates the measured benefit of predictive representations from the physiological prior used to organize their training.
\end{abstract}


\section{Introduction}
\label{sec:intro}

\begin{figure}[ht!]
\centering
\includegraphics[width=0.99\linewidth]{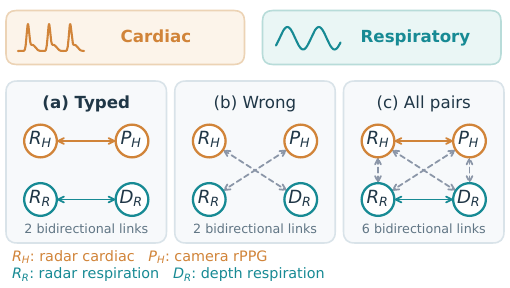}
\vspace{-4mm}
\caption{Candidate prediction graphs. Cardiac streams are orange, respiratory teal. Typed edges pair measurements of the same phenomenon, the wrong graph crosses phenomena with the same number of directed predictions, and all pairs has six links and more predictor parameters. Every link trains in both directions.}
\label{fig:graphs}
\end{figure}

Millimeter-wave radar and RGB-D cameras recover cardiac and respiratory waveforms without electrodes, which makes continuous monitoring practical in homes, clinics, and vehicles. Such systems accumulate synchronized signals at scale, whereas annotation follows a different economy because protocol labels describe short scripted episodes and require supervised acquisition sessions. The practical problem is therefore a large volume of unlabeled multimodal physiology in addition to a small volume of labeled episodes.

Representation learning addresses this imbalance by pretraining on unlabeled signals before a small supervised stage. For physiological time series the dominant family is contrastive learning, which requires two design decisions. The first is a set of transformations that must preserve the downstream class. The second is a rule that declares which window pairs are negative. Neither decision is settled for cardiorespiratory signals, because time dilation alters breathing rate, amplitude scaling alters a volume proxy, and two windows separated in time can describe the same physiological state~\cite{Iglesias2023Data}. Joint-embedding predictive architectures replace negative-pair discrimination with prediction of stopped target embeddings~\cite{assran2023ijepa}, and synchronous streams can supply these views without synthetic signal transformations.

A predictive objective still leaves a structural choice that is specific to multimodal physiology. When a recording contains more than one generator, cardiac and respiratory activity in this case, the objective must decide which streams predict one another. Previous multimodal predictive models align all available channels or learn the relationship from the data~\cite{dutta2026charm}. Instead, we encode physiological knowledge directly as a typed graph whose edges join streams intended to measure the same generator, as shown in Figure~\ref{fig:graphs}. This choice is testable rather than assumed, because a wrong-edge graph with identical capacity provides a matched control, and an all-pairs graph provides a denser alternative.

We study that question on contactless cardiorespiratory sensing and report both a positive and a negative outcome. Predictive pretraining improves label efficiency against a matched supervised model, whereas the typed graph does not establish an advantage over its capacity-matched and denser controls. We therefore treat the physiological prior as a stated hypothesis with explicit controls, rather than as an assumed source of the gain. The contributions are as follows:

\begin{enumerate}
\itemsep0.05em
\item Phenomenon-Graph JEPA maps four processed univariate streams to a joint representation through paired temporal and band-limited spectral branches, typed cross-sensor prediction and within-episode forward prediction without negative pairs, with optional Takens-inspired delay coordinates.
\item We measure label efficiency in the published OMuSense-23 split~\cite{canellas2026omusense}, with the complete recipe fixed in the validation data, and report paired inference at the participant-level under two configurations on the same ten test participants.
\item We test the typing assumption against matched wrong-edge and all-pairs graphs, and analyze which architectural components contribute to the measured gain.
\item We bound the result to wrist-only wearable physiology under two test protocols, with task-specific validation studies.
\end{enumerate}

Section~\ref{sec:related} reviews related work, Section~\ref{sec:method} formalizes the model and its objective, Section~\ref{sec:setup} describes datasets, protocol and statistical inference, Section~\ref{sec:results} reports the results, and Section~\ref{sec:conclusion} concludes.

\section{Related Work}
\label{sec:related}

\subsection{Contrastive pretraining for physiological signals}
Temporal and contextual contrasting learns biosignal representations from unlabeled data for downstream classification~\cite{eldele2021tstcc}. In OMuSense-23, a multimodal framework applies this approach to four processed cardiorespiratory streams~\cite{canellas2025ssl}. A later study combines contrastively pretrained convolutional features with handcrafted descriptors through cross attention~\cite{canellas2025fusion}. The contrastive pretraining stage uses augmented views and in-batch negatives, while augmentation effects vary between tasks~\cite{Iglesias2023Data}. Reviews of augmentation for time series report that the same transformation can help one task and damage another, which makes the choice part of the experimental hypothesis rather than a generic invariance and motivates alternatives with fewer transformation assumptions.

\subsection{Predictive architectures for time series}
JEPA-style learning predicts latent representations without negative pairs, with applications to generic time series~\cite{ennadir2025tsjepa}, multi-horizon forward prediction~\cite{lee2026cfjepa}, electrocardiography~\cite{weimann2025ecgjepa}, intensive-care physiology~\cite{fox2026physiojepa}, contactless radar~\cite{liu2026mmjepa} and time--frequency alignment~\cite{chaykowsky2026tfjepa}. Constant embeddings can minimize a cosine prediction loss, because a target that never varies is perfectly predictable. Variance and covariance regularization provide explicit constraints against collapse and redundant features~\cite{bardes2022vicreg}, and we adopt this regularization in Section~\ref{sec:objective}.

\subsection{Structure across channels}
CHARM learns inter-channel relations in heterogeneous time series~\cite{dutta2026charm}, while CardioState-JEPA learns a shared cardiac representation from electrocardiography, photoplethysmography and phonocardiography with explicit delay alignment~\cite{shafiq2026cardiostate}. Both treat the relation between channels as something to be learned or assumed. We examine whether restricting prediction to streams assigned to the same physiological phenomenon improves representations when cardiac and respiratory activity coexist. Answering this question requires a control that changes only the graph topology. A wrong-edge control with identical predictor capacity isolates the contribution of physiological edge typing in contactless recognition.

\section{Methodology}
\label{sec:method}

\begin{figure*}[ht!]
\centering
\includegraphics[width=0.99\linewidth]{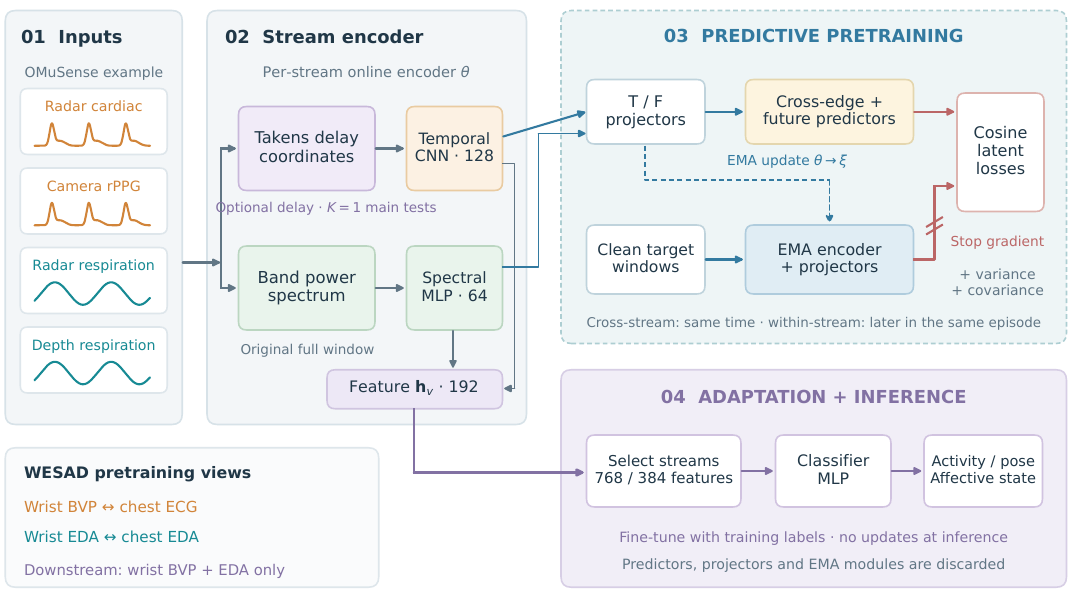}
\vspace{-3mm}
\caption{Phenomenon-Graph JEPA. Delay coordinates affect only the temporal input and are optional ($K=1$ in main tests). During pretraining, cross-edge and within-stream future predictors match stopped targets from EMA encoders and projectors. Recognition fine-tunes online encoders and a classifier, with no updates at inference. WESAD discards chest encoders after pretraining.}
\label{fig:arch}
\end{figure*}

The complete proposed model is depicted in Figure~\ref{fig:arch}. Each stream is encoded by a temporal and a spectral branch, pretraining predicts stopped target embeddings along the edges of a graph and forward in time, and the predictive modules are discarded before a classifier is fine-tuned on the labeled episodes. The following subsections define each component.

\subsection{Problem formulation and notation}
Let $V$ denote the set of processed streams. A window of stream $v\in V$ is a vector $\mathbf{x}_v\in\mathbb{R}^{L}$, whose length is given by $L=f_sT_w$ for sampling rate $f_s$ and duration $T_w$. This gives $L=200$ samples for OMuSense-23 ($T_w=10$~s, $f_s=20$~Hz) and $L=1{,}920$ for WESAD ($T_w=30$~s, $f_s=64$~Hz). A multi-stream window $\mathbf{X}\in\mathbb{R}^{|V|\times L}$ stacks the windows of all streams as rows. Throughout, $[\mathbf{a};\mathbf{b}]$ denotes the concatenation of two vectors. Every window belongs to one participant and to one episode, a contiguous interval recorded in a single annotated state. The pretraining observes the unlabeled windows of the training participants. Fine-tuning observes a labeled subset of complete training episodes, whose size is controlled by a requested label budget. The goal is to have an encoder $f_\theta:\mathbb{R}^{|V|\times L}\rightarrow\mathbb{R}^{192|V|}$ whose features make a classifier accurate with few labeled episodes.

\subsection{Typed phenomenon graph}
For a processed stream $v$ assigned to phenomenon $\phi(v)\in\{H,R\}$, cardiac or respiratory, we assume the shared-cause model expressed as:
\begin{equation}
X_v(t)=h_v\!\left(S_{\phi(v)}(t),U_v(t)\right)+\epsilon_v(t),
\label{eq:shared}
\end{equation}
where $S_\phi$ is latent physiology, $U_v$ is stream-specific state, $h_v$ contains acquisition and signal processing, and $\epsilon_v$ is noise. If two stream-specific terms are approximately conditionally independent given $S_\phi$, the predictable component between those streams should favour the shared generator. Equation~\eqref{eq:shared} states an inductive bias and not an identifiability result, because posture, motion, filtering, task order and cardiorespiratory coupling remain shared across streams.

The four nodes are radar cardiac $R_H$, remote photoplethysmography $P_H$, radar respiration $R_R$ and depth respiration $D_R$. The proposed edge set is given by:
\begin{equation}
E_\star=\{(R_H,P_H),\,(R_R,D_R)\},
\label{eq:graph}
\end{equation}
and the controls described in Figure~\ref{fig:graphs} are the wrong graph $E_{\mathrm{w}}=\{(R_H,D_R),(R_R,P_H)\}$, which crosses the phenomena while keeping the numbers of edges and directed predictors identical, and the complete graph $E_{\mathrm{all}}$ with all six edges. For each edge set $E\in\{E_\star,E_{\mathrm{w}},E_{\mathrm{all}}\}$, every undirected edge is trained in both directions, so the directed edge set is expressed as $\overrightarrow{E}=\{(u,v),(v,u):(u,v)\in E\}$. The typed and wrong graphs therefore have four directed edge predictors each, and all pairs have twelve. The typed and wrong graphs both train 2.19~million pretraining parameters, so their comparison changes topology alone, while all pairs trains 3.19~million and confounds topology with predictor capacity. The graph defines which predictions are trained. It is not a graph neural network, and no messages are exchanged at inference. The raw radar phase is excluded because it mixes motion with both phenomena.

\subsection{Paired temporal and spectral encoding}
Each stream is encoded twice. The temporal branch is a strided convolutional network given by:
\begin{equation}
\mathbf{g}_{T,v}(\mathbf{x}_v)=\mathbf{W}_T\,\mathrm{AvgPool}\!\left(\Phi_4\circ\Phi_3\circ\Phi_2\circ\Phi_1(\mathbf{x}_v)\right),
\label{eq:temporal}
\end{equation}
where each block $\Phi_i$ is a convolution, batch normalization and GELU activation with widths 64, 128, 128 and 128, kernels 7, 5, 5 and 5, and stride 2, and the linear map $\mathbf{W}_T$ gives a 128-dimensional output. The total stride is 16 samples, and adaptive average pooling allows the same architecture to serve both window lengths.

The spectral branch receives the original full window. After removing the mean, a least-squares linear trend is subtracted, giving $\tilde{x}_v[n]$, and a Hann taper $w[n]$ is applied. The power spectrum is expressed as:
\begin{equation}
P_v[k]=\frac{\left|\sum_{n=0}^{L-1}w[n]\,\tilde{x}_v[n]\,e^{-\mathrm{j}2\pi kn/L}\right|^2}{\sum_{n=0}^{L-1}w[n]^2},
\label{eq:power}
\end{equation}
at frequencies $f_k=kf_s/L$. The branch keeps only the bins $\mathcal{K}_{\phi}=\{k: f_k\in[f_{\phi}^{\mathrm{lo}},f_{\phi}^{\mathrm{hi}}]\}$ of the assigned phenomenon, which are 0.8 to 4~Hz for cardiac nodes and 0.1 to 0.6~Hz for respiratory nodes. With the shorthand $\mathcal{K}=\mathcal{K}_{\phi(v)}$, the spectral descriptor combines the normalized spectral shape with the total power of the band and is given by:
\begin{equation}
\mathbf{s}_v=\bigg[\Big(\log\frac{P_v[k]}{\sum_{k'\in\mathcal{K}}P_v[k']}\Big)_{k\in\mathcal{K}};\,\log\sum_{k\in\mathcal{K}}P_v[k]\bigg],
\label{eq:spectral_feature}
\end{equation}
whose first part is the normalized spectral shape and whose last entry is the total band power. On OMuSense-23 this yields 33 cardiac and 6 respiratory bins, hence input dimensions of 34 and 7, and on WESAD 97 cardiac and 15 electrodermal bins, hence 98 and 16. A layer normalization and a multilayer perceptron with 128 hidden units map $\mathbf{s}_v$ to a 64-dimensional feature $\mathbf{g}_{F,v}\in\mathbb{R}^{64}$. The downstream and stream representations are then expressed as:
\begin{equation}
\begin{aligned}
\mathbf{h}_v&=\big[\mathbf{g}_{T,v};\,\mathbf{g}_{F,v}\big]\in\mathbb{R}^{192},\\
\mathbf{h}&=\big[\mathbf{h}_{v_1};\dots;\mathbf{h}_{v_{|V|}}\big]\in\mathbb{R}^{192|V|},
\end{aligned}
\label{eq:representation}
\end{equation}
which gives 768 features for the four OMuSense streams, without a learned fusion layer. The spectral branch makes the rate and harmonic structure explicit, on which the tasks depend, while the temporal branch keeps event order and morphology. The band restriction separates the phenomena at the input, and Section~\ref{sec:components} reports its measured contribution separately from the predictive objective.

As an optional input organization we also test Takens-inspired delay coordinates~\cite{takens1981attractors}, which stack $K$ lagged copies of the signal before the temporal branch only. The delay vector is given by:
\begin{equation}
\mathbf{y}_v[n]=\big[x_v[n],\,x_v[n-\tau_v],\,\dots,\,x_v[n-(K-1)\tau_v]\big],
\label{eq:delay}
\end{equation}
for $n\geq (K-1)\tau_v$, so only indices with a complete history are used and no fabricated history enters the encoder. The lag $\tau_v$ follows the phenomenon of the node, and the spectrum is still computed from the original full window. A convolutional filter already reads lagged samples, so the informative control is not the shorter unmodified encoder but a learned front end with the same temporal support. This learned-history control is a convolution from one input channel to $K$ channels with kernel length $(K-1)\tau_v+1$, followed by the same temporal encoder. All main OMuSense and WESAD test configurations use $K=1$. The $K=3$ validation studies use cardiac and respiratory lags of 2 and 10 samples (0.10 and 0.50~s) for short histories and 4 and 20 samples (0.20 and 1.00~s) for long histories. This construction does not establish the assumptions of a dynamical reconstruction theorem for noisy physiological recordings.

\subsection{Predictive objective}
\label{sec:objective}
Each stream has temporal and spectral projectors, producing $\mathbf{z}^{T,v}_t$ and $\mathbf{z}^{F,v}_t$ from the two encoder features at time $t$. Exponential moving average (EMA) copies of both encoders and projectors produce stopped targets $\bar{\mathbf{z}}^{T,v}_t$ and $\bar{\mathbf{z}}^{F,v}_t$, and the target parameters $\xi$ follow the online parameters $\theta$ through the update given by:
\begin{equation}
\xi\leftarrow\mu\,\xi+(1-\mu)\,\theta,\qquad \mu=0.996.
\label{eq:ema}
\end{equation}
Let $\mathbf{c}^v_t=[\mathbf{z}^{T,v}_t;\mathbf{z}^{F,v}_t]$ denote the combined online projection. Temporal edge prediction uses this combined source to predict a temporal target, a separate spectral predictor maps between spectral projections, and a within-stream predictor forecasts a later window of the same episode. The three prediction losses are expressed as:
\begin{align}
\Lcross &=\frac{1}{|\overrightarrow{E}|}\sum_{(u,v)\in\overrightarrow{E}}
d\!\left(\psi^{u\to v}_{T}(\mathbf{c}^u_t),\sg(\bar{\mathbf{z}}^{T,v}_t)\right),
\label{eq:cross}\\
\Lspec &=\frac{1}{|\overrightarrow{E}|}\sum_{(u,v)\in\overrightarrow{E}}
d\!\left(\psi^{u\to v}_{F}(\mathbf{z}^{F,u}_t),\sg(\bar{\mathbf{z}}^{F,v}_t)\right),
\label{eq:spectral}\\
\Lfut &=\frac{1}{|V|}\sum_{v\in V}
d\!\left(\psi^{v}_{\mathrm{fut}}(\mathbf{c}^v_t,\mathbf{e}(\Delta)),\sg(\bar{\mathbf{z}}^{T,v}_{t+\Delta})\right),
\label{eq:future}
\end{align}
where $\psi^{u\to v}_{T}$, $\psi^{u\to v}_{F}$ and $\psi^{v}_{\mathrm{fut}}$ are online predictors with parameters in $\theta$, $\sg(\cdot)$ stops gradients, and $d$ is the cosine distance, given by:
\begin{equation}
d(\mathbf{a},\mathbf{b})=1-\frac{\mathbf{a}^{\top}\mathbf{b}}{\|\mathbf{a}\|_2\,\|\mathbf{b}\|_2}.
\label{eq:cosine}
\end{equation}
The displacement $\Delta$ in seconds between the context and the future window is encoded sinusoidally as:
\begin{equation}
\mathbf{e}(\Delta)=\Big[\big(\sin\omega_i\Delta\big)_{i=0}^{15};\,\big(\cos\omega_i\Delta\big)_{i=0}^{15}\Big]\in\mathbb{R}^{32},
\label{eq:time_code}
\end{equation}
with frequencies $\omega_i=100^{-i/16}$ and start displacements of 10 to 20~s on OMuSense-23 and 30 to 120~s on WESAD. Cross-edge pairs are taken at the same time $t$. Targets never provide inputs to the predictors, and both edge directions train all online stream encoders. Future windows remain within the same original state episode, do not overlap the context and never cross an acquisition gap. The loss uses no class targets, but annotation-derived episode boundaries constrain sampling, making the paired windows same-class by construction.

Because constant embedding minimizes every cosine term, variance and covariance penalties act on the online projections~\cite{bardes2022vicreg}. For a batch of $N$ centered projections $\mathbf{Z}\in\mathbb{R}^{N\times D}$ with columns $\mathbf{Z}_{:,j}$ and covariance matrix $\mathbf{C}=\mathbf{Z}^{\top}\mathbf{Z}/(N-1)$, they are given by:
\begin{align}
\Lvar&=\frac{1}{D}\sum_{j=1}^{D}\max\!\Big(0,\,1-\sqrt{\mathrm{Var}(\mathbf{Z}_{:,j})+10^{-4}}\Big),\label{eq:var}\\
\Lcov&=\frac{1}{D}\sum_{j\neq j'}C_{jj'}^{2},\label{eq:cov}
\end{align}
averaged over the temporal and spectral projections. The variance term keeps every dimension active, and the covariance term prevents dimensions from duplicating one another. The complete objective is expressed as:
\begin{equation}
\mathcal{L}=\Lcross+0.3\Lspec+\Lfut+0.1\Lvar+0.004\Lcov,
\label{eq:objective}
\end{equation}
with coefficients compared on validation data before each configuration was fixed for its evaluation.

\subsection{Downstream adaptation}
\label{sec:adaptation}
After pretraining, predictors, projectors and EMA modules are discarded. A stream-availability vector $\mathbf{a}\in\{0,1\}^{|V|}$ marks the observed streams, and absent streams are set to zero before encoding, so that the classifier input is expressed as:
\begin{equation}
\mathbf{r}=\big[f_\theta\big(\operatorname{diag}(\mathbf{a})\,\mathbf{X}\big);\,\mathbf{a}\big],
\label{eq:classifier_input}
\end{equation}
where $\operatorname{diag}(\mathbf{a})$ multiplies each stream, a row of $\mathbf{X}$, by its availability flag. The input width is therefore 772 on OMuSense-23 and 386 for the two wrist streams of WESAD. The classifier has one hidden layer of 128 GELU units with dropout 0.2, and it is trained with cross-entropy. The main comparisons fine-tune both the encoder and the classifier in the training participants, with learning rates of $10^{-4}$ and $10^{-3}$, respectively, and inference does not perform parameter or normalization updates.

The second configuration adds a bounded temporal shift during downstream training. A single integer lag $\ell\sim\mathcal{U}\{-10,\dots,10\}$ is drawn per sample and shared across streams, and the shifted window is given by:
\begin{equation}
x^{(\ell)}_v[n]=x_v\big[\min(\max(n-\ell,0),L-1)\big],
\label{eq:shift}
\end{equation}
which moves events by at most 0.5~s without changing their spacing, so breathing and cardiac rate are preserved. Edges are replicated rather than zero-filled, which avoids an identifiable discontinuity, and a shared lag preserves the cross-sensor alignment that the graph objective predicts across. The same configuration samples sensor subsets for $\mathbf{a}$ during downstream training.

\section{Experimental Evaluation}
\label{sec:setup}

\subsection{Datasets}
\label{sec:datasets}
\noindent\textbf{OMuSense-23} includes 50 participants recorded with millimeter-wave radar and RGB-D cameras in supine, seated and standing poses during four instructed breathing activities~\cite{canellas2026omusense}. We used four published processed streams at 20~Hz, radar cardiac, rPPG, radar respiration, and depth respiration. Removing idle intervals and center cropping or edge padding nominal 30~s episodes without rate warping yields 600 episodes and 12,600 ten-second windows at a 1~s step. Tasks are breathing activity (four classes) and pose (three). Channel statistics are fitted on training episodes before overlapping windows are formed and are fixed thereafter for every split and branch.

\noindent\textbf{WESAD} includes 15 participants wearing chest RespiBAN and wrist Empatica E4 devices during baseline, stress, amusement and guided meditation~\cite{schmidt2018wesad}. Pretraining pairs wrist blood volume pulse (BVP, an optical pulse waveform) with chest electrocardiography (ECG) for the cardiac phenomenon, and wrist with chest electrodermal activity (EDA, skin conductance) for the electrodermal phenomenon. The cardiac and EDA bands are 0.8--4 and 0.02--0.5~Hz, respectively. Signals are resampled to 64~Hz in state-pure 30~s windows. \textbf{Only wrist BVP and EDA are used downstream.} Both chest encoders are removed before classifier training, fine-tuning and testing, leaving 384 features. Thus, chest data provide privileged pretraining views from training participants only. Models are trained in WESAD without transfer from OMuSense-23 using training-only normalization.

\subsection{Evaluation protocol}
\label{sec:protocol}
\begin{figure*}[ht!]
\centering
\includegraphics[width=\textwidth]{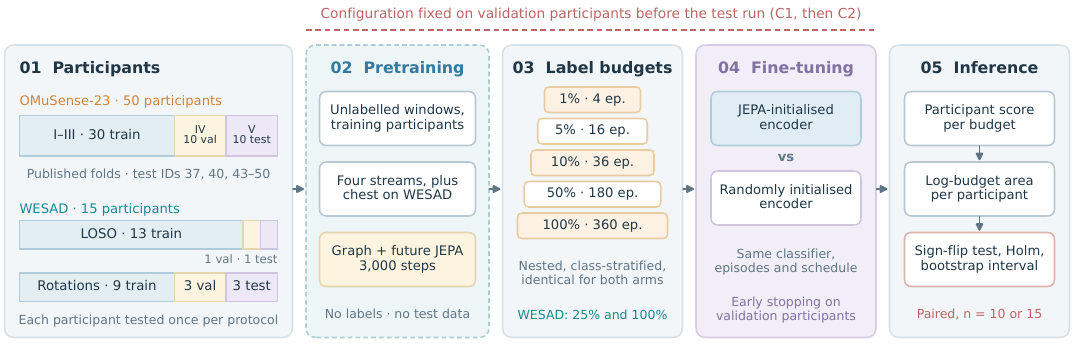}
\vspace{-7mm}
\caption{Evaluation protocol. Participants are separated before pretraining, which uses unlabeled windows of training participants only. Nested label budgets are identical for the JEPA-initialized and randomly initialized arms, early stopping uses validation participants, and each configuration is fixed before its test run. Inference operates on participant-level areas.}
\label{fig:protocol}
\vspace{-3mm}
\end{figure*}

The evaluation follows a freeze-then-test protocol, as depicted in Figure~\ref{fig:protocol}. OMuSense-23 uses the published folds, where folds I to III provide 30 training participants, fold IV provides ten validation participants and fold V provides ten test participants, with identifiers 37, 40 and 43 to 50. These participants also appeared in earlier project evaluations, so subsequent fixed configurations are not independent-cohort replications. Pretraining uses only training participants, which is stricter than the four non-test folds used in~\cite{canellas2025ssl}. Each design decision was made on the validation participants, and each configuration was fixed before its test run.

Labeled subsets are nested by episode and class-stratified, and they are identical across methods for each budget and seed. For a requested budget $b$, the achieved fraction is given by:
\vspace{-1mm}
\begin{equation}
\rho_b=\frac{|\mathcal{B}_b|}{|\mathcal{B}_{\mathrm{full}}|},\qquad \mathcal{B}_{1}\subset\mathcal{B}_{2}\subset\dots\subset\mathcal{B}_{\mathrm{full}},
\label{eq:budget}
\end{equation}
where $\mathcal{B}_b$ is the set of labeled training episodes. On OMuSense-23 the requested 1, 5, 10, 50 and 100\% select 4, 16, 36, 180 and 360 episodes, which correspond to achieved fractions of 1.11, 4.44, 10, 50 and 100\% and to 2 to 180 minutes of labeled signal. WESAD requests 25 and 100\%, and the lower budget selects 6 of 27 rotation episodes (22.22\%) or 9 of 39 LOSO episodes (23.08\%). Labels used for checkpoint selection on validation participants are additional to the reported budgets, and analyzes use achieved rather than requested fractions.

On WESAD the principal task is meditation versus baseline. Leave-one-subject-out (LOSO) testing uses 13 training, one validation and one test participant with a 0.25~s step, following the outer split and windowing of~\cite{alvarez2021meditation}, but reserving a validation participant for neural early stopping and using fewer modalities. Five rotations use 9, 3 and 3 participants and a 5~s step. Each participant is tested once per protocol. Exploratory one-seed runs reuse the rotation validation participants for stress versus non-stress (baseline plus amusement), three-class baseline/stress/amusement, and four classes including meditation. The first two follow the reference task definitions, while the four-class task is an extension. These scores also use checkpoint-selection participants, so they are reported separately from test results.

\subsection{Configurations and baselines}
\label{sec:configurations}
The base configuration \textbf{C1} uses the temporal--spectral encoder and the objective of Equation~\eqref{eq:objective}. Configuration \textbf{C2} adds the bounded shift of Equation~\eqref{eq:shift} and sensor-subset masking to both JEPA and supervision, retaining the encoder architecture and pretraining objective, so it changes downstream training only. C1 was fixed before its test, and C2 was fixed afterwards on validation data and evaluated on the same test participants. Supervised controls train identical backbones and classifiers from random initialization on identical episodes with the same schedule.

The \textbf{prior contrastive SSL reimplementation}~\cite{canellas2025ssl} shares the four streams, train-fold normalization, test cohort and episode budgets, but probes frozen 512-dimensional features with a linear classifier over five seeds, whereas the present comparisons fine-tune 768-dimensional features over three seeds. It uses generic in-batch negatives without the published negative-window separation of more than 20~s and assumes some architecture details, so its curve is a historical system comparison, not an exact replication or an objective-only contrast. Published fusion~\cite{canellas2025fusion} reports fixed-split and five-rotation scores, which are not interchangeable with our fold-V results. WESAD references use handcrafted EDA, acceleration, temperature and BVP-derived heart features rather than two learned streams.

\subsection{Metrics and statistical inference}
\label{sec:metrics}
Per-participant scores are macro-F1 on OMuSense-23 and balanced accuracy (BA) on WESAD, following its reference. For $Q$ classes indexed by $c$, with true positives $\mathrm{TP}_c$, false positives $\mathrm{FP}_c$ and false negatives $\mathrm{FN}_c$, the two scores are given by:
\begin{align}
\mathrm{F1}&=\frac{1}{Q}\sum_{c=1}^{Q}\frac{2\,\mathrm{TP}_c}{2\,\mathrm{TP}_c+\mathrm{FP}_c+\mathrm{FN}_c},\label{eq:f1}\\
\mathrm{BA}&=\frac{1}{Q}\sum_{c=1}^{Q}\frac{\mathrm{TP}_c}{\mathrm{TP}_c+\mathrm{FN}_c}.\label{eq:ba}
\end{align}
The label-efficiency endpoint summaries the score curve of participant $i$ over $B$ budgets in a single number. The achieved fractions are mapped to a normalized logarithmic axis, $\lambda_b=(\log\rho_b-\log\rho_1)/(\log\rho_B-\log\rho_1)$, so that $\lambda_1=0$ and $\lambda_B=1$, and the area is expressed as the trapezoidal sum:
\vspace{-1mm}
\begin{equation}
A_i=\sum_{b=1}^{B-1}\left(\lambda_{b+1}-\lambda_b\right)\frac{S_{i,b}+S_{i,b+1}}{2},
\label{eq:area}
\end{equation}
where $S_{i,b}$ is the score at budget $b$ after averaging seeds within the participant. The area lies in the score range and is not a receiver operating characteristic area. With the two WESAD budgets it reduces to the mean of the two endpoint scores.

For two methods and $n$ test participants, the paired difference of participant $i$ is $\delta_i=A_i^{(1)}-A_i^{(2)}$, and the reported effect is the mean difference $\bar\delta=n^{-1}\sum_{i=1}^{n}\delta_i$ in percentage points. Significance uses an exact paired sign-flip permutation test over all $2^n$ sign patterns, assuming sign exchangeability of paired differences under the null, and the two-sided probability is given by:
\begin{equation}
p=\frac{1}{2^{n}}\sum_{\boldsymbol{\eta}\in\{-1,1\}^{n}}\mathbb{I}\!\left(\Big|\sum_{i=1}^{n}\eta_i\delta_i\Big|\geq\Big|\sum_{i=1}^{n}\delta_i\Big|\right),
\label{eq:signflip}
\end{equation}
where $\boldsymbol{\eta}$ is a vector of signs and $\mathbb{I}(\cdot)$ is the indicator function, equal to one when its argument holds and zero otherwise. With ten participants who all favour one method, only the identity and the global sign reversal reach the observed statistic, which gives $p=2/1024\approx0.0020$. Paired $t$ and Wilcoxon tests give consistent conclusions. Because each configuration declares more than one comparison, Holm correction controls the family-wise error within the three C1 contrasts, the two C2 contrasts and each single-contrast WESAD family, excluding other exploratory analyses. For the $m$ ordered raw values $p_{(1)}\leq\dots\leq p_{(m)}$ of a family, the adjusted values are expressed as
\begin{equation}
\tilde{p}_{(k)}=\max_{l\leq k}\,\min\!\big\{1,\,(m-l+1)\,p_{(l)}\big\}.
\label{eq:holm}
\end{equation}
Intervals are percentile intervals from 50{,}000 participant bootstrap resamples of $\bar\delta$. Overlapping windows and repeated seeds do not increase $n$, and overlapping LOSO training cohorts make WESAD uncertainty approximate.

\subsection{Implementation details}
\label{sec:implementation}
All settings are listed in Table~\ref{tab:impl}. Pretraining uses 3{,}000 steps with a batch size of 64, AdamW with weight decay $10^{-4}$ and a cosine-annealed learning rate of $10^{-3}$. Downstream training uses validation-based early stopping within 40 epochs and a patience of six epochs on OMuSense-23 and the WESAD rotations. The WESAD LOSO campaign uses at most ten epochs and a patience of three, a computational choice that we disclose because LOSO requires fifteen folds per seed. OMuSense-23 tests use three seeds, WESAD tests two and validation studies one. C1 and the WESAD fits use no stochastic augmentation of the pretraining context. Experiments run locally on an RTX 2080 SUPER and on the Roihu cluster.

\begin{table}[ht!]
\def\arraystretch{1.05}
\setlength{\tabcolsep}{1.85em}
\centering
\caption{Implementation settings shared by all main comparisons. Loss weights refer to the cross-edge, spectral, future, variance and covariance terms of Equation~\eqref{eq:objective}.}
\label{tab:impl}
\vspace{3pt}
\footnotesize
\begin{tabular}{@{}ll@{}}
\toprule
Setting & Value \\
\midrule
Temporal CNN & widths 64/128/128/128, kernels 7/5/5/5 \\
Branch and stream widths & 128 temporal, 64 spectral, 192 \\
Loss weights & 1, 0.3, 1, 0.1, 0.004 \\
Pretraining & 3{,}000 steps, batch 64, AdamW \\
Pretraining rate & $10^{-3}$, cosine, weight decay $10^{-4}$ \\
EMA momentum & 0.996 \\
Classifier & 128 GELU units, dropout 0.2 \\
Downstream rates & head $10^{-3}$, encoder $10^{-4}$ \\
Epochs / patience & 40/6, WESAD LOSO 10/3 \\
C2 shift bound & $\pm10$ samples ($\pm0.5$ s) \\
Seeds & 3 OMuSense, 2 WESAD, 1 validation \\
\bottomrule
\end{tabular}

\end{table}

\section{Results}
\label{sec:results}

\subsection{Label efficiency}
\begin{figure}[ht!]
\centering
\includegraphics[width=0.99\linewidth]{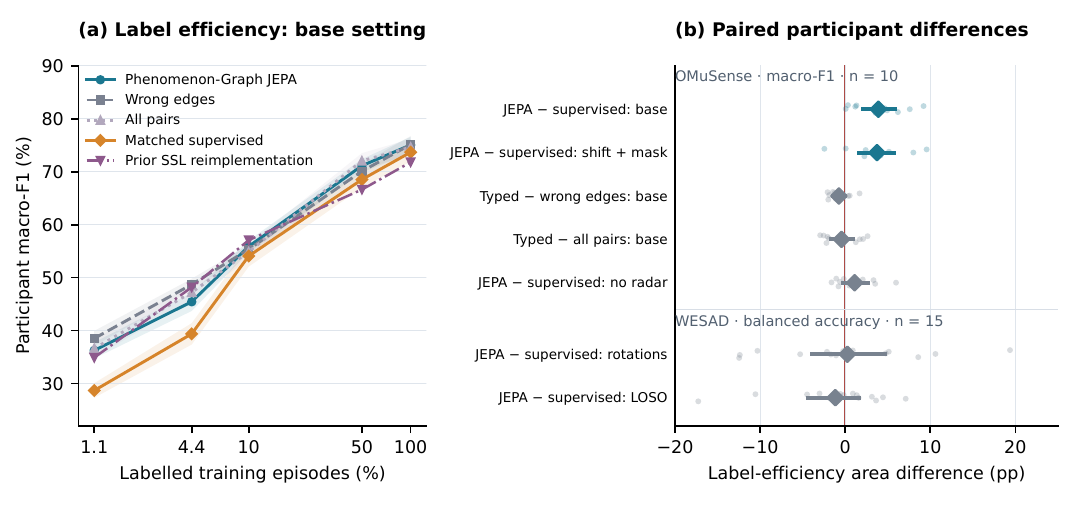}
\vspace{-3mm}
\caption{(a) C1 activity macro-F1 with wrong-edge, all-pairs and matched supervised controls. Shading is the participant standard error after three-seed averaging. Prior SSL is a five-seed historical frozen-probe reimplementation~\cite{canellas2025ssl}, not a matched control. (b) The seven declared contrasts, with participant points, mean diamonds and 95\% bootstrap intervals.}
\label{fig:results}
\end{figure}

C1 improves the area over the matched supervision by 3.91 points at Holm $p=0.006$ (95\% interval 2.08 to 5.80), favoring JEPA for all ten participants, as shown in Table~\ref{tab:stats}. Macro-F1 is 36.3 versus 28.7 on nominal 1\% labels and 75.1 versus 73.7 on full labels, as shown in Figure~\ref{fig:results}(a). C2 gives 3.74 points for the same people, 39.9 versus 27.4 at 1\%, but 76.5 versus 78.6 at full labels, so its downstream transformations help both systems, with supervision leading at full labels.

\vspace{-2mm}
\begin{table}[ht!]
\def\arraystretch{1.05}
\setlength{\tabcolsep}{0.5em}
\centering
\caption{Participant-level differences in normalized log-budget area (pp). Positive values favour the first method, Holm correction applies within each declared family as described in Section~\ref{sec:metrics}, and intervals are 95\% participant bootstrap intervals.}
\label{tab:stats}
\vspace{3pt}
\footnotesize
\begin{tabular}{@{}lrcrr@{}}
\toprule
Comparison & $\Delta$ & 95\% interval & Raw $p$ & Holm $p$ \\
\midrule
\multicolumn{5}{@{}l}{\textit{OMuSense-23, fold-V test, macro-F1 area, $n=10$}} \\
JEPA $-$ supervised, C1 & $\mathbf{+3.91}$ & $[+2.08, +5.80]$ & 0.0020 & $\mathbf{0.0059}$ \\
JEPA $-$ supervised, C2 & $\mathbf{+3.74}$ & $[+1.68, +5.78]$ & 0.0117 & $\mathbf{0.0234}$ \\
Typed $-$ wrong edges, C1 & $-0.74$ & $[-1.44, +0.03]$ & 0.1016 & 0.2031 \\
Typed $-$ all pairs, C1 & $-0.43$ & $[-1.65, +0.87]$ & 0.5352 & 0.5352 \\
JEPA $-$ sup., C2, no radar & $+1.12$ & $[-0.25, +2.65]$ & 0.1992 & 0.1992 \\
\multicolumn{5}{@{}l}{\textit{WESAD, wrist-only test, BA area, $n=15$}} \\
JEPA $-$ sup., rotations & $+0.25$ & $[-3.88, +4.62]$ & 0.9129 & 0.9129 \\
JEPA $-$ sup., LOSO & $-1.16$ & $[-4.39, +1.56]$ & 0.5054 & 0.5054 \\
\bottomrule
\end{tabular}

\end{table}

The complete curves behind these areas are listed in Table~\ref{tab:curves}. Under C1, pretraining is ahead of supervision in all five budgets, and the difference is largest in the lowest budgets, 7.6 points at 1.11\% and 6.1 points at 4.44\%, before narrowing to 1.4 points with full labels. Under C2, pretraining is ahead at 1.11, 4.44 and 10\%, whereas supervision is ahead at 50 and 100\%. The claim therefore concerns label efficiency, measured by the area, and not a universal improvement at full supervision.

\subsection{Comparison with prior work on this corpus}
Published fusion remains stronger, as shown in Table~\ref{tab:comparison}. Its fixed-split scores match Fold I in its five-fold table~\cite{canellas2025fusion}, which also reports 56.40 for sex and 79.20 for joint pose--activity recognition, different endpoints, so we keep those aggregates separate from our fold-V evaluation. Because the fixed split cannot be identified with fold V from the printed fold names alone, we do not treat the two as the same held-out cohort. The reimplementation reaches 72.71 activity accuracy against 74.00 reported originally, but preprocessing, adaptation, seed counts and aggregation prevent objective-only comparisons. At nominal 1\% labels, JEPA and supervised accuracy are 38.25 versus 33.78 in C1 and 40.95 versus 34.14 in C2. Accuracy and balanced accuracy coincide on these class-balanced participant windows, whereas macro-F1 differs, so the accuracy values in this comparison are not interchangeable with the macro-F1 values listed in Table~\ref{tab:curves}.

\begin{table}[ht!]
\def\arraystretch{1.05}
\setlength{\tabcolsep}{0.9em}
\centering
\caption{OMuSense-23 activity macro-F1 (\%) on the fold-V test participants at each achieved label fraction, as means over ten participants after averaging seeds. The historical reimplementation uses five seeds and a frozen linear probe.}
\label{tab:curves}
\vspace{3pt}
\footnotesize
\begin{tabular}{@{}lrrrrr@{}}
\toprule
Achieved labels (\%) & 1.11 & 4.44 & 10 & 50 & 100 \\
\midrule
\multicolumn{6}{@{}l}{\textit{C1, base configuration}} \\
Supervised & 28.68 & 39.38 & 54.09 & 68.53 & 73.70 \\
JEPA, typed graph & 36.25 & 45.44 & 55.91 & 71.14 & 75.07 \\
JEPA, wrong edges & 38.52 & 48.72 & 55.36 & 70.11 & 75.15 \\
JEPA, all pairs & 36.63 & 47.23 & 54.90 & 72.07 & 74.64 \\
\multicolumn{6}{@{}l}{\textit{C2, shift + sensor masking}} \\
Supervised & 27.36 & 41.89 & 52.96 & 72.60 & 78.62 \\
JEPA, typed graph & 39.92 & 45.91 & 57.51 & 71.67 & 76.53 \\
\multicolumn{6}{@{}l}{\textit{Historical reimplementation, frozen probe}} \\
Prior contrastive SSL & 34.90 & 48.19 & 57.09 & 66.64 & 71.81 \\
\bottomrule
\end{tabular}

\end{table}

The macro-F1 curve of the historical reimplementation is listed in Table~\ref{tab:curves}. It is below C1 pretraining at 1.11, 50 and 100\%, but above it at 4.44 and 10\%. Because its adaptation, representation width and seed count differ, these crossings describe two systems and not a controlled comparison of objectives.

\begin{table}[ht!]
\def\arraystretch{1.05}
\setlength{\tabcolsep}{2.6em}
\centering
\caption{Full-label OMuSense accuracy (\%), four streams. Published fixed-split and five-fold results are separate aggregates, and the reimplementation differs in adaptation and seed count. Current rows are matched within C1 or C2 on fold V.}
\label{tab:comparison}
\vspace{3pt}
\footnotesize
\begin{tabular}{@{}lrr@{}}
\toprule
Method & Activity & Pose \\
\midrule
\multicolumn{3}{@{}l}{\textit{Published, fixed split, four streams}} \\
Contrastive SSL~\cite{canellas2025ssl} & 74.00 & 78.00 \\
Handcrafted + XGBoost~\cite{canellas2025fusion} & 81.00 & 86.00 \\
Handcrafted + deep fusion~\cite{canellas2025fusion} & 85.00 & 97.00 \\
\multicolumn{3}{@{}l}{\textit{Published, five-fold average}} \\
Handcrafted + deep fusion~\cite{canellas2025fusion} & 87.20 & 92.60 \\
\multicolumn{3}{@{}l}{\textit{Reimplementation, fold V, frozen probe, 5 seeds}} \\
Prior contrastive SSL & 72.71 & --- \\
\multicolumn{3}{@{}l}{\textit{This work, fold V, fine-tuned, 3 seeds}} \\
Supervised, base (C1) & 73.98 & 80.87 \\
Phenomenon-Graph JEPA, C1 & 75.52 & 82.41 \\
Supervised, shift + masking (C2) & 78.93 & 84.15 \\
Phenomenon-Graph JEPA, C2 & 76.94 & 84.78 \\
\bottomrule
\end{tabular}

\end{table}

\subsection{Recognition across tasks and protocols}
Table~\ref{tab:tasks} separates both completed WESAD meditation tests from the exploratory task studies. At full labels LOSO slightly favors JEPA and rotations are nearly tied, with large participant variation, and accuracy falls as classes rise from two to four. The one-seed scores are exploratory rather than held-out, because those participants also determine early stopping, and the published references use different classifiers and features. On OMuSense-23 the C2 pose accuracy is 84.78 against 84.15, with a smaller participant spread for JEPA.

\begin{table}[ht!]
\def\arraystretch{1.05}
\setlength{\tabcolsep}{2.3em}
\centering
\caption{Balanced accuracy (\%), mean $\pm$ SD across participant seed means. Groups separate test from exploratory validation scores. Published 30~s LOSO references~\cite{alvarez2021meditation}, which add handcrafted wrist-feature families, are $^{a}$82.80$\pm$6.30, $^{b}$92.43$\pm$11.04 and $^{c}$73.95$\pm$16.78, contextual rather than matched. Stress versus rest is stress against baseline plus amusement, three states are baseline/stress/amusement, and four states add meditation.}
\label{tab:tasks}
\vspace{3pt}
\footnotesize
\begin{tabular}{@{}lrr@{}}
\toprule
Labels & Supervised & JEPA \\
\midrule
\multicolumn{3}{@{}l}{\textit{OMuSense-23, C2, fold-V test, three seeds}} \\
Pose, 3 classes & $84.15\pm10.29$ & $\mathbf{84.78}\pm7.29$ \\
Activity, 4 classes & $\mathbf{78.93}\pm3.65$ & $76.94\pm3.35$ \\
\multicolumn{3}{@{}l}{\textit{WESAD meditation vs baseline$^{a}$, LOSO, two seeds}} \\
25\% & $\mathbf{62.14}\pm13.14$ & $58.36\pm13.15$ \\
100\% & $70.45\pm17.72$ & $\mathbf{71.91}\pm19.10$ \\
\multicolumn{3}{@{}l}{\textit{WESAD meditation vs baseline$^{a}$, five rotations}} \\
25\% & $59.72\pm9.24$ & $\mathbf{60.45}\pm11.86$ \\
100\% & $\mathbf{71.00}\pm14.89$ & $70.78\pm14.45$ \\
\multicolumn{3}{@{}l}{\textit{WESAD exploratory validation rotations, one seed}} \\
Stress vs rest$^{b}$ & $87.40\pm10.67$ & $\mathbf{88.10}\pm8.89$ \\
Three states$^{c}$ & $63.61\pm9.61$ & $\mathbf{65.72}\pm14.75$ \\
Four states & $\mathbf{55.94}\pm10.96$ & $54.73\pm9.45$ \\
\bottomrule
\end{tabular}

\end{table}

\subsection{No established advantage from physiological typing}
The wrong-edge graph described in Figure~\ref{fig:graphs}(b) matches the typed graph in edge count, directed predictor count and 2.19~million pretraining parameters, so the comparison changes topology alone. Neither control separates from the typed graph, as shown in Table~\ref{tab:stats}, and the three graph curves stay within 3.3 macro-F1 points in every budget, the largest separation favoring a falsifier, as listed in Table~\ref{tab:curves}. The comparisons establish a pretraining benefit over supervision but not that the prescribed pairing explains it, and non-significance is not equivalence, because no equivalence margin was declared. Shared task timing and cardiorespiratory coupling can make a wrong edge predictive, while the band-limited spectral branch may already separate the phenomena at the input. These explanations remain unresolved.

The validation evidence was not a reliable guide for this question. At the 10\% budget the typed graph led the wrong graph by approximately 3.4 macro-F1 points on the validation participants, and this difference decreased to 0.55 points on the test participants. The same evaluation does not contain an outer-test model with every cross-sensor prediction term removed, so the controls determine which of the evaluated pairings is preferable, and they do not attribute the pretraining benefit to cross-sensor prediction as such.

\subsection{Component analysis}
\label{sec:components}
Spectral inputs raise the supervised macro-F1 from 49.5 to 52.5 at 10\% labels and 71.1 to 74.3 at full labels, while the spectral prediction weight of 0.3 versus zero gives $+2.37/-0.41$ on those budgets, a positive two-budget area difference of 0.98 points. Increasing the spectral weight from 0.3 to 1.0 reduces this area by 1.63 points, which shows that a larger coefficient did not help, not that the spectral target is useless. Noise and attention do not help their controls, as summarized in Table~\ref{tab:ablation}, whereas downstream shifts add 1.27 and 1.54 area points to JEPA and supervision. The token-predictor comparison matches target tokens and approximately matches parameter counts, but the attention implementation encodes source-token times while the mean-pooling predictor ignores them, so it is not a one-factor intervention on attention alone.

\begin{table}[ht!]
\def\arraystretch{1.05}
\setlength{\tabcolsep}{2.3em}
\centering
\caption{OMuSense validation, one seed. Area differences (pp), each against its own reference condition on the listed grid, so grids are not interchangeable. Attention controls match capacity but differ in source-time inputs. Validation estimates, not significance tests.}
\label{tab:ablation}
\vspace{3pt}
\footnotesize
\begin{tabular}{@{}lcr@{}}
\toprule
Within-study comparison & Budgets (\%) & $\Delta$ area \\
\midrule
Spectral weight, 1.0 vs 0.3 & 10/100 & $-1.63$ \\
Additive noise, pretraining & 1/10/100 & $-1.60$ \\
Additive noise, downstream & 1/10/100 & $-0.69$ \\
Bounded shift, downstream & 10/100 & $+1.27$ \\
Cross-attention vs token MLP & 1/10/100 & $-1.70$ \\
Attention vs MLP classifier & 1/10/100 & $-1.43$ \\
Rate $3{\times}10^{-4}$ vs $10^{-3}$ & 1/10/100 & $+0.63$ \\
Short delay vs no delay & 10/100 & $+1.02$ \\
Short delay vs learned history & 10/100 & $+1.44$ \\
Long delay vs learned history & 10/100 & $+0.09$ \\
\bottomrule
\end{tabular}

\end{table}

Short delays help against both no delay and matched learned history, on one seed only, and main tests exclude them. The endpoint values of this study are listed in Table~\ref{tab:delay}. Short delay coordinates reduce the 10\% endpoint relative to no delay and improve the full-label endpoint, so the gain in area comes from the higher budget and is not an established improvement in the lowest-label regime. The long-delay condition matches its learned-history control to within 0.09 area points. Sensor masking improves radar-removal validation by 12.9 points, while its test contrast is inconclusive. C2 changes the shift and availability training together, so the C2 test result does not independently confirm the effect of the shift.

\begin{table}[ht!]
\def\arraystretch{1.05}
\setlength{\tabcolsep}{1.8em}
\centering
\caption{Delay-coordinate validation study on fold IV, one seed, fine-tuned activity macro-F1 (\%) at 10\% and 100\% labels and their normalized log-budget area. Short and long lags are 2/10 and 4/20 samples for cardiac/respiratory nodes at 20~Hz.}
\label{tab:delay}
\vspace{3pt}
\footnotesize
\begin{tabular}{@{}lccc@{}}
\toprule
Input organisation & 10\% & 100\% & Area \\
\midrule
No delay & 59.45 & 71.91 & 65.68 \\
Short delay coordinates & 58.56 & \textbf{74.83} & \textbf{66.70} \\
Short learned-history control & 56.93 & 73.58 & 65.26 \\
Long delay coordinates & 58.85 & 72.47 & 65.66 \\
Long learned-history control & 58.67 & 72.46 & 65.57 \\
\bottomrule
\end{tabular}

\end{table}

\subsection{Boundary on wearable physiology}
Neither wrist-only WESAD test establishes the advantage seen on OMuSense, at $+0.25$ points for rotations and $-1.16$ for LOSO. The published 30~s reference, $82.80\pm6.30$ from LDA with EDA, acceleration, temperature and heart features~\cite{alvarez2021meditation}, remains higher, but the gap does not isolate a feature or training effect. Two-budget area merely averages endpoints, and task, cohort, sensors and schedules prevent attributing the corpus difference to contact sensing alone. Corrected wrist-only graph controls were not run, so the WESAD evaluations do not decide the typing hypothesis on wearable physiology.

\subsection{Limitations}
Four limitations bound the interpretation. First, all OMuSense comparisons use the same ten test participants, which also appeared in earlier project evaluations, and C2 was fixed after C1 had been evaluated, so the two positive results are observations under two configurations on one cohort rather than independent replications. Second, within-episode sampling uses annotation-derived episode boundaries, so the pretraining pipeline is free of class targets but not of annotation information. Third, the Holm families do not correct every exploratory comparison made across the complete project. Fourth, the WESAD and OMuSense outcomes differ in task, budgets, participants, window lengths and training schedules, so their contrast does not isolate sensor contact as a causal factor.

\section{Conclusion}
\label{sec:conclusion}
We introduced Phenomenon-Graph JEPA, a representation-learning framework combining temporal--spectral encoding, typed cross-sensor prediction and within-episode forward prediction. On OMuSense-23, pretraining improved normalised label-efficiency area over matched supervised training by 3.91 and 3.74 percentage points under two configurations evaluated on the same test cohort. However, comparisons with wrong-edge and all-pairs graphs did not establish an advantage of physiological edge typing. The observed gains therefore support the predictive learning framework under the evaluated conditions, without establishing that the proposed graph structure accounts for them. Short Takens-inspired delay coordinates improved validation performance relative to a learned-history control with matched temporal support, but were not included in the main test evaluations. Neither wrist-only WESAD protocol established a corresponding label-efficiency advantage. These findings distinguish the measured benefit on OMuSense-23 from the unresolved contributions of physiological typing and delay coordinates. Further evaluation should compare against stronger feature-fusion baselines under matched protocols and assess generalisation to independent cohorts.

\section*{\centering\normalsize ACKNOWLEDGMENT}
The research was supported by the Research Council of Finland Smart Video Sensorization for Secure Healthcare Monitoring (SViSenS) project (grant 370277), the Interreg Aurora ResilientEdge project (grant 20373282) and the Profi7 Hybrid Intelligence programme (352788). The authors acknowledge the use of AI-assisted tools for language editing, with a focus on grammar checking and readability, and for coding support. The Elicit platform was used during the literature search as an AI-assisted discovery aid for candidate records. All inclusion decisions, exclusions, data extraction, synthesis, interpretation and final writing decisions were performed by the authors.

\section*{\centering\normalsize COMPLIANCE WITH ETHICAL STANDARDS}
This work is a secondary analysis of two existing human-subject corpora, OMuSense-23~\cite{canellas2026omusense} and WESAD~\cite{schmidt2018wesad}. No new data were collected and no participant intervention was performed. The OMuSense-23 source reports written informed consent and adherence to the Declaration of Helsinki. No attempt was made to re-identify participants.

\balance
\bibliographystyle{IEEEbib}
\bibliography{refs_v4b}

@inproceedings{alvarez2021meditation,
  title     = {Meditation Detection Using Sensors from Wearable Devices},
  author    = {{\'{A}lvarez Casado}, Constantino and Paananen, Petteri and Siirtola, Pekka and Pirttikangas, Susanna and {Bordallo L\'{o}pez}, Miguel},
  booktitle = {Adjunct Proceedings of the 2021 ACM International Joint Conference on Pervasive and Ubiquitous Computing and the 2021 ACM International Symposium on Wearable Computers},
  pages     = {112--116},
  publisher = {ACM},
  year      = {2021},
  doi       = {10.1145/3460418.3479318},
  url       = {https://doi.org/10.1145/3460418.3479318}
}

@inproceedings{schmidt2018wesad,
  title     = {Introducing {WESAD}, a Multimodal Dataset for Wearable Stress and Affect Detection},
  author    = {Schmidt, Philip and Reiss, Attila and Duerichen, Robert and Marberger, Claus and Van Laerhoven, Kristof},
  booktitle = {Proceedings of the 20th ACM International Conference on Multimodal Interaction},
  pages     = {400--408},
  publisher = {ACM},
  year      = {2018},
  doi       = {10.1145/3242969.3242985},
  url       = {https://doi.org/10.1145/3242969.3242985}
}

@article{canellas2026omusense,
  title   = {{OMuSense-23}: A Multimodal Dataset for Contactless Breathing Pattern Recognition and Biometric Analysis},
  author  = {{Lage Ca\~{n}ellas}, Manuel and Nguyen, Le and Mukherjee, Anirban and {\'{A}lvarez Casado}, Constantino and Wu, Xiaoting and Nguyen, Nhi and Susarla, Praneeth and Sharifipour, Sasan and Jayagopi, Dinesh B. and {Bordallo L\'{o}pez}, Miguel},
  journal = {IEEE Internet of Things Journal},
  pages   = {1--14},
  year    = {2026},
  note    = {Early access},
  doi     = {10.1109/JIOT.2026.3720896},
  url     = {https://doi.org/10.1109/JIOT.2026.3720896}
}

@article{canellas2025ssl,
  title   = {A Self-Supervised Multimodal Framework for 1D Physiological Data Fusion in Remote Health Monitoring},
  author  = {{Lage Ca\~{n}ellas}, Manuel and {\'{A}lvarez Casado}, Constantino and Nguyen, Le and {Bordallo L\'{o}pez}, Miguel},
  journal = {Information Fusion},
  volume  = {124},
  pages   = {103397},
  year    = {2025},
  doi     = {10.1016/j.inffus.2025.103397},
  url     = {https://doi.org/10.1016/j.inffus.2025.103397}
}

@inproceedings{canellas2025fusion,
  title     = {Fusion of Handcrafted and Deep-Learned Cardiorespiratory Features for Breathing Pattern Classification},
  author    = {{Lage Ca\~{n}ellas}, Manuel and {\'{A}lvarez Casado}, Constantino and Malin, Miika and Prencipe, Nicoletta and Sharifipour, Sasan and {Bordallo L\'{o}pez}, Miguel},
  booktitle = {Proceedings of the 33rd European Signal Processing Conference (EUSIPCO)},
  pages     = {1627--1631},
  year      = {2025}
}

@inproceedings{eldele2021tstcc,
  title     = {Time-Series Representation Learning via Temporal and Contextual Contrasting},
  author    = {Eldele, Emadeldeen and Ragab, Mohamed and Chen, Zhenghua and Wu, Min and Kwoh, Chee Keong and Li, Xiaoli and Guan, Cuntai},
  booktitle = {Proceedings of the Thirtieth International Joint Conference on Artificial Intelligence},
  pages     = {2352--2359},
  year      = {2021},
  doi       = {10.24963/ijcai.2021/324},
  url       = {https://doi.org/10.24963/ijcai.2021/324}
}

@inproceedings{assran2023ijepa,
  title     = {Self-Supervised Learning from Images with a Joint-Embedding Predictive Architecture},
  author    = {Assran, Mahmoud and Duval, Quentin and Misra, Ishan and Bojanowski, Piotr and Vincent, Pascal and Rabbat, Michael and LeCun, Yann and Ballas, Nicolas},
  booktitle = {Proceedings of the IEEE/CVF Conference on Computer Vision and Pattern Recognition},
  pages     = {15619--15629},
  year      = {2023},
  url       = {https://openaccess.thecvf.com/content/CVPR2023/html/Assran_Self-Supervised_Learning_From_Images_With_a_Joint-Embedding_Predictive_Architecture_CVPR_2023_paper.html}
}

@inproceedings{bardes2022vicreg,
  title     = {{VICReg}: Variance-Invariance-Covariance Regularization for Self-Supervised Learning},
  author    = {Bardes, Adrien and Ponce, Jean and LeCun, Yann},
  booktitle = {International Conference on Learning Representations},
  year      = {2022},
  url       = {https://openreview.net/forum?id=xm6YD62D1Ub}
}

@article{weimann2025ecgjepa,
  title   = {Self-Supervised Pre-Training with Joint-Embedding Predictive Architecture Boosts {ECG} Classification Performance},
  author  = {Weimann, Kuba and Conrad, Tim O. F.},
  journal = {Computers in Biology and Medicine},
  volume  = {196},
  number  = {Part B},
  pages   = {110809},
  year    = {2025},
  doi     = {10.1016/j.compbiomed.2025.110809},
  url     = {https://doi.org/10.1016/j.compbiomed.2025.110809}
}

@article{ennadir2025tsjepa,
  title         = {Joint Embeddings Go Temporal},
  author        = {Ennadir, Sofiane and Golkar, Siavash and Sarra, Leopoldo},
  journal       = {arXiv preprint arXiv:2509.25449},
  year          = {2025},
  eprint        = {2509.25449},
  archivePrefix = {arXiv},
  url           = {https://arxiv.org/abs/2509.25449}
}

@inproceedings{chaykowsky2026tfjepa,
  title     = {{TF-JEPA}: Predictive Alignment of Time--Frequency Representations Without Contrastive Pairs},
  author    = {Chaykowsky, Michael},
  booktitle = {1st ICLR Workshop on Time Series in the Age of Large Models},
  year      = {2026},
  url       = {https://iclr.cc/virtual/2026/10013864}
}

@article{lee2026cfjepa,
  title         = {{CF-JEPA}: Mask-Free Forward Prediction with Asymmetric Encoder Utilization for Time-Series Representation Learning},
  author        = {Lee, Jaehoon and Sim, Sunghyun},
  journal       = {arXiv preprint arXiv:2606.07031},
  year          = {2026},
  eprint        = {2606.07031},
  archivePrefix = {arXiv},
  url           = {https://arxiv.org/abs/2606.07031}
}

@inproceedings{fox2026physiojepa,
  title     = {{PhysioJEPA}: Joint Embedding Representations of Physiological Signals for Real Time Risk Estimation in the Intensive Care Unit},
  author    = {Fox, Benjamin and Hoang, Dung and Jiang, Joy and Jayaraman, Pushkala and Parekh, Ankit and Nadkarni, Girish N. and Sakhuja, Ankit},
  booktitle = {Proceedings of the Fifth Machine Learning for Health Symposium},
  series    = {Proceedings of Machine Learning Research},
  volume    = {297},
  pages     = {120--135},
  year      = {2026},
  publisher = {PMLR},
  url       = {https://proceedings.mlr.press/v297/fox26a.html}
}

@inproceedings{liu2026mmjepa,
  title     = {{mmJEPA-ECG}: Cross-Posture Robust Contactless Electrocardiogram Monitoring via Millimeter Wave Radar Sensing},
  author    = {Liu, Ziyang and He, Siyuan and Liang, Feng and Huang, Chang and Zhong, Shuxin and Wu, Kaishun},
  booktitle = {Proceedings of the AAAI Conference on Artificial Intelligence},
  volume    = {40},
  pages     = {38962--38970},
  year      = {2026},
  doi       = {10.1609/aaai.v40i45.41242},
  url       = {https://doi.org/10.1609/aaai.v40i45.41242}
}

@article{shafiq2026cardiostate,
  title         = {{CardioState-JEPA}: Delay-Aware Cross-Modal Learning of a Shared Cardiac Representation},
  author        = {Shafiq, Hamza and Pham, Hung Manh and Zhu, Bin and Zhou, Pan and Hu, Jun and Saeed, Aaqib},
  journal       = {arXiv preprint arXiv:2608.12944},
  year          = {2026},
  eprint        = {2608.12944},
  archivePrefix = {arXiv},
  url           = {https://arxiv.org/abs/2608.12944}
}

@inproceedings{dutta2026charm,
  title     = {Giving Sensors a Voice: Multimodal {JEPA} for Semantic Time-Series Embeddings},
  author    = {Dutta, Utsav and Pastrana, Gerardo and Pakazad, Sina Khoshfetrat and Ohlsson, Henrik},
  booktitle = {Proceedings of the 43rd International Conference on Machine Learning},
  series    = {Proceedings of Machine Learning Research},
  volume    = {306},
  year      = {2026},
  note      = {In press},
  eprint    = {2605.31580},
  archivePrefix = {arXiv},
  url       = {https://arxiv.org/abs/2605.31580}
}

@article{Iglesias2023Data,
	author = {Iglesias, Guillermo and Talavera, Edgar and Gonz{\' a}lez-Prieto, {\' A}ngel and Mozo, Alberto and G{\' o}mez-Canaval, Sandra},
	journal = {Neural Computing and Applications},
	doi = {10.1007/s00521-023-08459-3},
	issn = {0941-0643},
	number = {14},
	year = {2023},
	month = {mar 24},
	pages = {10123--10145},
	publisher = {{Springer Science and Business Media LLC}},
	title = {Data {Augmentation} techniques in time series domain: a survey and taxonomy},
	url = {http://dx.doi.org/10.1007/s00521-023-08459-3},
	volume = {35},
}

@incollection{takens1981attractors,
  title={Detecting Strange Attractors in Turbulence},
  author={Takens, Floris},
  booktitle={Dynamical Systems and Turbulence, Warwick 1980},
  editor={Rand, David and Young, Lai-Sang},
  series={Lecture Notes in Mathematics}, volume={898}, pages={366--381},
  publisher={Springer}, year={1981}, doi={10.1007/BFb0091924}
}

\end{document}